\documentclass[manuscript,screen,review=false]{acmart}

\AtBeginDocument{%
  \providecommand\BibTeX{{%
    \normalfont B\kern-0.5em{\scshape i\kern-0.25em b}\kern-0.8em\TeX}}}

\setcopyright{acmcopyright}
\copyrightyear{2018}
\acmYear{2018}
\acmDOI{XXXXXXX.XXXXXXX}

\usepackage{amsmath}
\usepackage{amsfonts}

\usepackage{graphicx}
\usepackage{subcaption}

\usepackage{pgfplots}
\pgfplotsset{compat=newest}

\usepackage{longtable}
\usepackage{multirow}
\usepackage{changepage}

\usepackage{csquotes}

\usepackage{listings}
\usepackage{color}
\definecolor{backcolor}{RGB}{250, 250, 250}
\definecolor{cmtcolor}{RGB}{140, 250, 140}
\definecolor{kwcolor}{RGB}{60, 150, 240}
\definecolor{numcolor}{RGB}{120, 120, 120}
\definecolor{strcolor}{RGB}{240, 210, 60}
\usepackage{cite}

\usepackage{comment}
\usepackage{xcolor}
\usepackage{soul}

\begin{document}

\title[Bio-Inspired Deep Learning]{Spiking Neural Networks and Bio-Inspired Supervised Deep Learning: A Survey}

\author{Gabriele Lagani}
\authornote{Corresp.}
\email{gabriele.lagani@isti.cnr.it}
\author{Fabrizio Falchi}
\email{fabrizio.falchi@isti.cnr.it}
\author{Claudio Gennaro}
\email{claudio.gennaro@isti.cnr.it}
\author{Giuseppe Amato}
\email{giuseppe.amato@isti.cnr.it}
\affiliation{%
  \institution{ISTI-CNR}
  \city{Pisa}
  \country{Italy}
  \postcode{56124}
}

\renewcommand{\shortauthors}{Lagani, et al.}

\begin{abstract}
  For a long time, biology and neuroscience fields have been a great source of inspiration for computer scientists, towards the development of Artificial Intelligence (AI) technologies. This survey aims at providing a comprehensive review of recent biologically-inspired approaches for AI. After introducing the main principles of computation and synaptic plasticity in biological neurons, we provide a thorough presentation of Spiking Neural Network (SNN) models, and we highlight the main challenges related to SNN training, where traditional backprop-based optimization is not directly applicable. Therefore, we discuss recent bio-inspired training methods, which pose themselves as alternatives to backprop, both for traditional and spiking networks. Bio-Inspired Deep Learning (BIDL) approaches towards advancing the computational capabilities and biological plausibility of current models. 
\end{abstract}



\received{}
\received[revised]{}
\received[accepted]{}

\maketitle

\section{Introduction}

Spiking Neural Networks (SNN) have recently emerged as a biologically inspired alternative to traditional Deep Learning (DL) models, towards addressing the current limitations of Deep Neural Networks (DNNs) in terms of ecological impact \citep{badar2021}. Indeed, biological brains exhibit extraordinary capabilities in terms of energy efficiency, supporting advanced cognitive functions while consuming only 20W \citep{javed2010}. It is believed that the key to the energy efficient computation of biological neurons lies in the particular coding paradigm based on short pulses, or \textit{spikes} \citep{gerstner}. SNN models aim at simulating the behavior of biological neurons more realistically, compared to traditional DNNs. As a result, SNNs are well suited for energy-efficient implementations in \textit{neuromorphic} \citep{shrestha2022, huynh2022, schuman2022, zhu2020, roy2019b} or \textit{biological} \citep{ruaro2005, lagani2021a, kagan2022} hardware. This makes SNNs a promising direction toward energy-efficient DL.

Unfortunately, training SNNs is not trivial, as traditional optimization based on the backpropagation algorithm (\textit{backprop}) is not directly applicable \citep{ponulak2005}. In fact, the biological plausibility of backprop -- the workhorse of DL -- is questioned by neuroscientists \citep{oreilly, marblestone2016, hassabis2017, lake2017, richards2019}. Therefore, researchers took again inspiration from biology, in order to find new learning solutions as alternatives to backprop. The goal was not only to address the problem of SNN training \citep{neftci2019, dampfhoffer2023}, but also to discover novel approaches to the learning problem \citep{scellier2017, millidge2020, hinton2022}, and possibly more data efficient strategies \citep{gupta2022, journe2022, lagani2021b, lagani2021c, lagani2022b, lagani2022c}. In fact, another limitation of current DL solutions is the requirement of large amounts of training data. On the other hand, biological brains also show interesting properties in terms of data efficiency, being able to infer new knowledge and easily generalize from little experience \citep{lake2020}. 

In this survey, we illustrate biologically realistic SNN models of neural computation, and we discuss spike coding strategies, spiking models of synaptic plasticity, and the training challenges and application potentials of this line of research. We also consider the learning problem in more detail, illustrating recently proposed methodologies for SNN and DNN training, which pose themselves as alternatives to traditional backprop-based training.

\begin{figure*}
    \centering
    \includegraphics[width=0.6\textwidth]{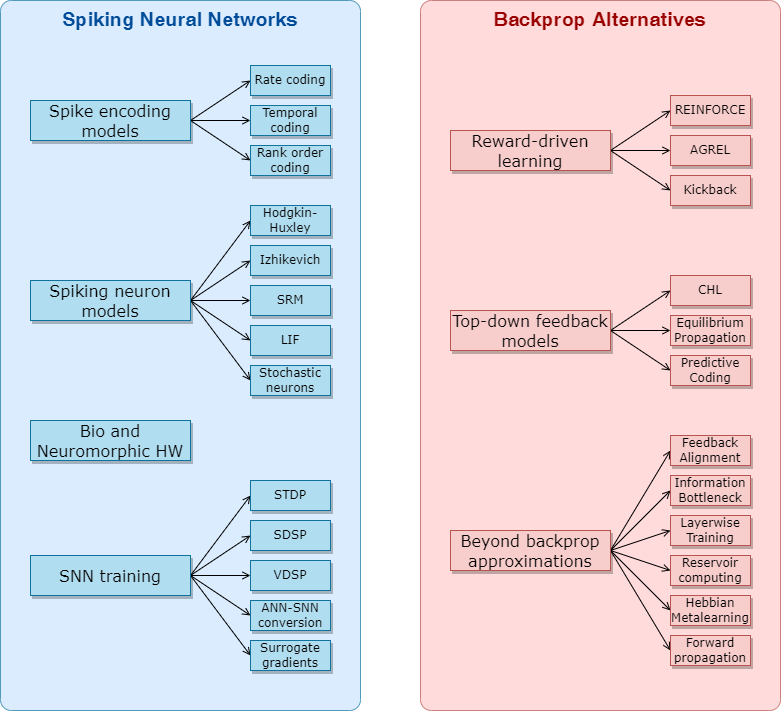}
    \caption{A schematic view of the topics of Bio-Inspired Deep Learning (BIDL) addressed in this work. 
    We discuss SNN models, which aim at providing a biologically faithful simulation of real neurons, and they find applications in the context of energy-efficient biological and neuromorphic computing. Learning strategies for SNNs include Spike Time Dependent Plasticity (STDP), surrogate gradient strategies for adapting backprop optimization to the case of SNNs, etc.
    We also discuss training algorithms that pose themselves as alternatives to backprop. Some approaches aim at approximating backprop using only biologically plausible synaptic updates, while others, rather than simply approximating backprop, consider different perspectives, such as forward propagation approaches, reservoir computing, etc.
    }
    \label{fig:bidl}
\end{figure*}
 
One of the goals of this survey is to provide a comprehensive review of the methods coming from biological inspiration, and their potential impact on current DL technologies. The field of Bio-Inspired Deep Learning (BIDL) observes the convergence of a broad spectrum of ideas, coming from the computer science and neuroscience fields; hence, one of the goals of this work is to highlight the connections between these two viewpoints. Fig. \ref{fig:bidl} provides a schematic visualization of the topics addressed in this work.

This document can be of interest both to novel readers who approach themes in the BIDL field for the first time, as well as a reference for researchers already familiar with these topics. Moreover, this document does not require prerequisite knowledge in the neuroscience domain, but the necessary background is provided where needed. At the same time, neuroscientists that are curious about the engineering aspects behind AI architectures could also find this document an interesting resource.

This survey is organized as follows:
\begin{itemize}
    \item Section \ref{sec:related} discusses related surveys in the BIDL field.
    \item Section \ref{sec:plasticity} briefly introduces some background about biological neurons and synaptic plasticity models.
    \item Section \ref{sec:spiking} discusses biologically detailed models of neural computation based on SNNs, highlights the challenges of training such models, but also their technological potential for energy-efficient biological and neuromorphic computing.
    \item Section \ref{sec:bp_alternatives} explores alternative DNN training algorithms which do not require backprop. 
    \item Finally, we present our conclusions in Section \ref{sec:concl}, outlining open challenges and possible future research directions.
\end{itemize}

Moreover, in our companion paper \citep{survey_plasticity} we discuss biological models of synaptic plasticity in greater detail, and we highlight the relationships between such models and unsupervised pattern discovery mechanisms in neural networks, showing the connections between neuroscientific principles and emerging cognitive behavior.

\section{Related Surveys} \label{sec:related}

The field of BIDL has been the subject of increasing attention recently, and it has been reviewed in a number of recent contributions from different perspectives. In particular, there are two perspectives from which the field is approached: the neuroscience and the computer science/engineering viewpoints. Related contributions so far have been more tied to one of these viewpoints, or they have tackled specific aspects of this field. In this perspective, the goal of our contribution is to provide a broader perspective on the various concepts that emerge in the field of BIDL, and to highlight the connections between the neuroscience viewpoint, and the computer science/engineering aspects.

A number of recent reviews \citep{marblestone2016, hassabis2017, richards2019, lake2017} address the field of BIDL from a high-level perspective, discussing the mutual benefits that exploration in biologically grounded mechanisms behind intelligence could bring both to the neuroscience and computer science fields. The authors also suggest possible specific directions of exploration where biological inspiration could play a crucial role.

A general background about biologically grounded neural system modeling can be found in the book from Gerstner and Kistler \citep{gerstner}, where a thorough introduction to SNN models is provided, as well as to Hebbian and Spike Time Dependent Plasticity (STDP) models.
A review more specifically focused on reward-modulated Hebbian learning approaches can be found in \citep{triche2022}, while recent SNN developments and applications are surveyed in \citep{pfeiffer2018, tavanaei2019a, nunes2022}. The discussions in \citep{neftci2019, snntorch} focus more specifically on backprop alternatives for SNNs. Finally, various recent surveys provide a thorough analysis of the field of neuromorphic hardware \citep{shrestha2022, huynh2022, schuman2022, zhu2020, roy2019b}.

Compared to previous surveys, we provide significant contributions in the following directions:
\begin{itemize}
    \item We emphasize the connections between low-level learning mechanisms and high-level abstractions, as opposed to works more focused on the high-level perspectives \citep{marblestone2016, hassabis2017, richards2019, lake2017};
    \item We develop a comprehensive viewpoint involving bio-inspired techniques ranging from synaptic plasticity models for spiking neurons, to supervised or reward-driven backprop alternatives, while other contributions only focus on backprop extensions for SNNs \citep{neftci2019, snntorch} or reward-driven approaches \citep{triche2022};
    \item We highlight the connections between traditional and spike-based models, showing the interplay between the two domains, and the potentials of SNN models for energy-efficient neuromorphic and biological computing, compared to works that focus more specifically on the low-level mechanisms of spike-based computation \citep{pfeiffer2018, tavanaei2019a, nunes2022}.
\end{itemize}

\section{Biological Neurons and Synaptic Plasticity} \label{sec:plasticity}

This Section is devoted to introducing the fundamental aspects of neuron biology, and the mechanisms of synaptic \textit{plasticity} underlying the learning behavior of biological brains. From this it will be possible to draw relationships between the computational learning mechanisms that we are going to discuss in the following, and the biological substrate that supports such mechanisms.

In the following, we start with an introduction on neural cell biology in subsection \ref{sec:plasticity:cells}, then we move to synaptic plasticity models based on the Hebbian principle in subsection \ref{sec:plasticity:hebbian}, highlighting the relationships between such models and emerging unsupervised pattern discovery mechanisms (such as clustering and subspace learning).

\subsection{Background on Neural Cells} \label{sec:plasticity:cells}

Neural cells are characterized by a central body, the \textit{soma}, which receives electric input stimuli through ramified connections called \textit{dendrites}, and emit signals through an output connection called \textit{axon} \citep{lytton, dayan, bear, white, sheperd}. The axon is connected to other neurons' dendrites by chemical couplings called \textit{synapses}. The strength of synaptic couplings is plastic, and its change determines the learning properties of neurons. The main aspects of plasticity are strengthening of synaptic efficacy, a.k.a. Long Term Potentiation (LTP), or weakening, a.k.a. Long Term Depression (LTD) \citep{bear1996, gerstner, white, sheperd}. Upon receiving input stimuli, an electric potential is accumulated on the neuronal membrane (i.e. the \textit{membrane potential}). When the membrane potential exceeds a threshold, an output stimulus is triggered, in terms of a pulse-shaped (a \textit{spike}) electric signal propagated on the axon \citep{gerstner}.

We can distinguish two main groups of neural cells: pyramidal and non-pyramidal \citep{white}.

Pyramidal cells represent the main computing unit in the brain. As the name suggest, they are characterized by a pyramid-like shape, from which two types of dendritic trees branch out: \textit{apical} and \textit{basal}. The former extends from the tip of the pyramid, and extend through cortical layers, while the latter extend mainly toward neighboring cells in the same region.
From the base of the pyramid, the axon of the neural cell originates. I then branches into a \textit{projection} axon, which can extend towards deeper cortical layers, and several \textit{axon collaterals}, which can either be \textit{local}, i.e. extending for a short distance towards neighboring neurons, or they can extend for longer distances either in the same layer or towards neighboring layers.

Non-pyramidal cells are instead a broad category comprising a variety of neural cell types \citep{stefanis2020}, but all sharing some common features such as the presence of a central body with a smaller size compared to pyramidal cells, the local connectivity patterns (i.e. short-distance), and the mainly inhibitory role of such neurons. A dense dendritic tree originates from the cell body, as well as an axon which tends to branch into multiple ramifications. Due to the local nature of non-pyramidal cell connectivity, which typically tend to target neighboring pyramidal cells, thus routing information in the local neighborhood, these neurons are also often referred to as \textit{interneurons}. Interneurons play an important role in inhibitory interaction and \textit{shunting inhibition} \citep{kubota2016} between neurons. The mechanisms of inhibitory \textit{lateral interaction}, mediated by non-pyramidal cells, are essential to achieve decorrelation in neural activity. In the following we will discuss how these mechanisms, coupled with appropriate synaptic plasticity models, allow neural circuits to implement unsupervised algorithms for automatic pattern discovery, thus showing an interesting mapping between biological circuits and learning mechanisms.

\subsection{Biological Models of Synaptic Plasticity} \label{sec:plasticity:hebbian}
\begin{figure}
    \centering
    \includegraphics[width=0.3\textwidth]{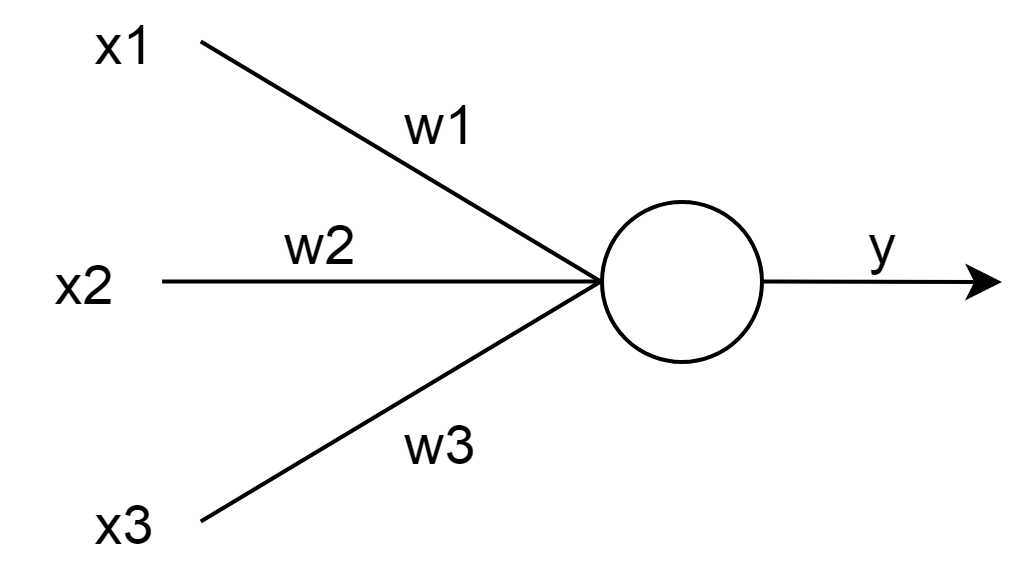}
    \caption{Representation of a neuron model with weights $w_1, w_2, w_3$, taking inputs $x_1, x_2, x_3$, and computing output $y = f(\sum_{i=1}^3 x_i w_i)$}
    \label{fig:neuron}
\end{figure}

In the following we will consider a neuron whose synaptic weights are described by a vector $\mathbf{w}$. The neuron takes as input a vector $\mathbf{x}$, and produces an output $y(\mathbf{x}, \mathbf{w}) = f(\mathbf{x}^T \, \mathbf{w})$ (Fig. \ref{fig:neuron}), where $f$ is the activation function (optionally, a bias term can be implicitly modeled as a synapse connected to a constant input). In the following, we use boldface fonts to denote vectors, and normal fonts to denote scalars. In neuroscientific terms, input values $\mathbf{x}$ are also termed \textit{pre-synaptic} activations, while the output $y$ is termed \textit{post-synaptic} activation. In order to draw a relationship between this abstract neuron model, and the biological neural cell structure described in the previous subsection, we can imagine that the output value of the abstract neuron model represents the \textit{rate} of output spikes generated by the neuron due to the input stimuli, hence we can talk about a \textit{rate-coded} neuron model \citep{gerstner}. Similarly, the input value on a given synapse corresponds to the rate of input spikes delivered through that synapse.

In order to model synaptic plasticity, neuroscientists propose the \textit{Hebbian} principle \citep{haykin, gerstner}: \textit{"fire together, wire together".} According to this principle, the synaptic coupling between two neurons is reinforced when the two neurons are simultaneously active \citep{haykin, gerstner}. Mathematically, this learning rule, in its simplest "vanilla" formulation, can be expressed as:
\begin{equation} \label{eq:hebbian_vanilla}
     \Delta w_i = \eta \, y \, x_i
\end{equation}
where $\eta$ is the learning rate, and the subscript $i$ refers to the i-th input/synapse.
The effect of this learning rule is essentially to consolidate correlated activations between neural inputs and outputs, by reinforcing the synaptic couplings, so that, if a similar input will be observed again in the future, a similar response will likely be elicited from the neuron.

The problem with Eq. \ref{eq:hebbian_vanilla} is that it is unstable, as repeated stimulation can induce the weights to grow unbounded. This issue can be counteracted by iteratively normalizing the weight vector \citep{oja1982}, or by adding a weight decay term $\gamma(\mathbf{w}, \mathbf{x})$ to the learning rule:
\begin{equation} \label{eq:hebbian_wd}
     \Delta w_i = \eta \, y \, x_i - \gamma(\mathbf{w}, \mathbf{x})
\end{equation}
In particular, with an appropriate choice for this term, i.e. $\gamma(\mathbf{w}, \mathbf{x}) = \eta \, y(\mathbf{w}, \mathbf{x}) \, w_i$, we obtain a learning rule that has been widely applied to the context of \textit{competitive learning} \citep{grossberg1976a, rumelhart1985, kohonen1982}:
\begin{equation} \label{eq:hebbian_centroid}
     \Delta w_i = \eta \, y \, x_i - \eta \, y \, w_i = \eta \, y \, (x_i - w_i)
\end{equation}

It can be shown that learning rule in Eq. \ref{eq:hebbian_vanilla} induces the weight vector of a neuron to align towards the principal component of the data distribution it is trained on \citep{oja1982, gerstner}, while Eq. \ref{eq:hebbian_centroid} pushes the weight vector towards the centroid of the cluster formed by the data points \citep{haykin, grossberg1976a}. This shows surprising connections between biological models of synaptic plasticity and computer science aspects related to data engineering and unsupervised pattern discovery mechanisms, specifically Principal Component Analysis (PCA) and clustering.

Note that, more in general, \textit{local} learning rules with higher order interactions can be considered. A general form for such a local synaptic update equation, for a generic synaptic connection $i$, can be expressed as \citep{gerstner}:
\begin{equation} \label{eq:hebbian_general}
     \Delta w_i = a_0 + a_1 x_i + a_2 y + a_3 x_i y + a_4 x_i^2 + a_5 y^2 + ...
\end{equation}
where the coefficients $a_i$ may depend on the weights.

When multiple neurons are involved in a complex network, it is desirable that different neurons learn to recognize different patterns. In other words, neural activity should pursue a \textit{decorrelated} form of coding \citep{foldiak1990, olshausen1996a}, in order to maximize the amount of information that neurons can transmit. This can be achieved by leveraging the biological mechanisms of \textit{lateral} and \textit{inhibitory} interaction. For example, Eq. \ref{eq:hebbian_centroid} can be coupled with \textit{competitive learning} mechanisms such as Winner-Takes-All (WTA) \citep{grossberg1976a, rumelhart1985}: in this case, the neuron with strongest similarity to a given input is chosen as the \textit{winner}, and it is the only one allowed to perform a weight update, by virtue of lateral inhibition \citep{gabbot1986}. In this way, if a similar input will be presented again in the future, the same neuron will be more likely to win again, and different neurons will specialize on different clusters. Other forms of lateral interaction can also enable a group of neurons to extract further principal components form the input (\textit{subspace learning}) \citep{sanger1989, becker1996a, karhunen1995}.

As a further extension of PCA-based learning methods, Independent Component Analysis (ICA) aims at removing higher-order correlations from data \citep{hyvarinen, haykin, amari1996, bell1995, pham1997, jutten1991}. Sparse Coding (SC) \citep{olshausen1996a} is another principle, strictly related to ICA \citep{olshausen1996b}, from which biologically plausible HaH feature learning networks can be derived, as in the Locally Competitive Algorithm (LCA) \citep{rozell2008}.

Hebbian approaches based on competitive \citep{lagani2019, lagani2021d, mthesis, krotov2019, wadhwa2016a, moraitis2021, miconi2021} or subspace learning \citep{lagani2022a, phdthesis, bahroun2017, illing2019} have recently been subject of significant interest towards applications in DL contexts.

\section{Spiking Neural Networks} \label{sec:spiking}

This Section discusses models of neural computation based on Spiking Neural Networks (SNNs) \citep{gerstner}, which more faithfully resemble real neurons compared to traditional ANN models. We start by introducing the various neuron models for SNN simulation. We highlight the applications related to biological and neuromorphic computing, which are of strong practical interest thanks to the energy efficiency of the underlying computing paradigm, and we discuss the challenges related to SNN traning. We describe the biological plasticity models for spiking neurons, and the approaches to translate backprop-based training to the spiking setting. Finally, an overview of some existing experimental results is given.

In the following, subsection \ref{sec:spiking:models} illustrates the main models of spiking neurons and spike-based strategies for information encoding; subsection \ref{sec:spiking:neuromorphic} highlights promising applications of spiking models in the field of neuromorphic and biological computing, and the challenges of learning with spiking models; therefore, subsection \ref{sec:spiking:plasticity} discusses the plasticity models for spiking neurons, based on Spike Time Dependent Plasticity (STDP); finally, subsection \ref{sec:spiking:dl} presents some experimental results from literature regarding the applications of SNNs and STDP in DL contexts.

\subsection{Spiking Neuron Models and Spike Codes} \label{sec:spiking:models}

Spiking Neural Networks (SNNs) are a realistic model of biological networks 
\citep{gerstner, maass1997}.
While in traditional Artificial Neural Networks (ANNs), neurons communicate via real-valued signals, in SNNs they emit short pulses called \textit{spikes}. 
All the spikes are equal to each other and values are encoded in the timing or in the frequency with which spikes are emitted. 

Various spiking neuron models have been proposed in literature \citep{gerstner}, which we highlight in the following:
\begin{itemize}
    \item Hodgkin-Huxley (HH) model \citep{hodgkin1952} is a classical and low-level description of the processes that regulate the neuron behavior, in terms of nonlinear differential equations. This model is very detailed, but also computationally expensive to simulate.
    \item Izhikevich model \citep{izhikevich2003} is less detailed but also computationally more efficient to simulate. It is based on a simplified description of the neuron behavior, compared to HH equations, but still retaining the main elements that allow to simulate a variety of observed neural behaviors, such as bursts of activity or sustained oscillations.
    \item The Spike Response Model (SRM) \citep{gerstner1995} is a higher-level description that describes the state of a neuron in terms of the electric potential accumulated on its neuronal membrane. Two kernels are defined which represent the effect of, respectively, incoming and outgoing spikes on the membrane potential. Appropriate choice of such kernels allows to simulate a variety of neuron behaviors.
    \item The Leaky Integrate and Fire (LIF) model \citep{abbott1993} is probably the most abstract description of spiking neurons, as well as the most computationally friendly for simulations. 
\end{itemize}

\begin{figure}
    \centering
    \includegraphics[width=0.4\textwidth]{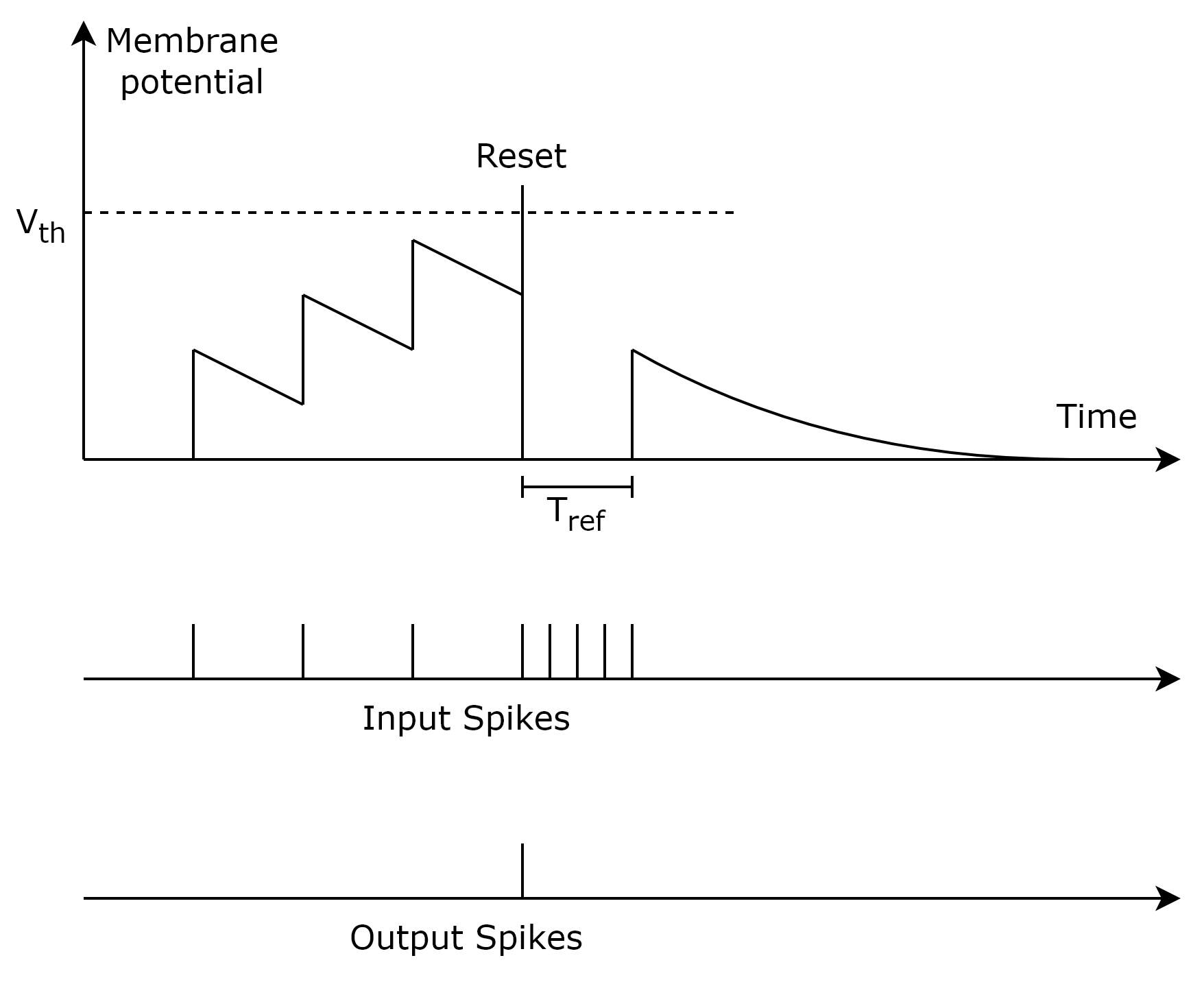}
    \caption{Behavior of a LIF unit.}
    \label{fig:lif}
\end{figure}

LIF neurons represent an ideal trade-off between biological accuracy and efficiency of simulation, when it comes to modeling complex spiking neural systems. For this reason, the LIF model is widely used in practice for the simulation of large-scale SNN architectures. In the following, we describe the LIF neuron model in more detail. LIF neurons behave like integrators, summing up all the received spikes (weighted by the synaptic coefficients) until a threshold is exceeded. At this point, an output spike is emitted (Fig. \ref{fig:lif}). In practice, this integration logic is implemented in terms of an electric potential that is accumulated on the neural membrane every time an input spike is received; when the threshold is reached and the output spike is released, the neural membrane discharges the accumulated potential and the process restarts. Immediately after the discharge, the neuron enters its \textit{refractory period}, a time interval where it cannot spike regardless of its input. These units are leaky in the sense that, when no spikes are received in input, the membrane potential decays exponentially.

Notice that the neural computation discussed above is deterministic, but extensions of neuron models to the stochastic setting also exist \citep{gerstner, gerstner1995}, which are better suited to model noisy scenarios. Specifically, there are two approaches to inject randomness in spiking models: \textit{noisy thresholds} (or \textit{escape noise}) and \textit{noisy integration} (or \textit{diffusive noise}). 
In noisy threshold models, the neuron doesn't fire deterministically when the threshold is overcome, but it fires stochastically, with a probability density that depends on the current value of the membrane potential itself (the higher the value, the more the neuron is likely to fire). Equivalently, the neuron fires when the threshold is overcome, but the value of the threshold is sampled randomly from a probability distribution at each time instant.
In noisy integration models, the neuron fires deterministically when the threshold is overcome, but integration is performed stochastically, i.e. adding a random noise contribution to the running sum at each time instant.

We mentioned at the beginning that information in spiking neurons is encoded in the timing or in the frequency of spikes. Let us elaborate further. Indeed, various models to encode information in spike sequences have been proposed \citep{auge2021, gerstner}, such as \textit{rate coding} (or \textit{frequency coding}), \textit{temporal coding}, and \textit{rank-order coding}.
In rate coding \citep{gerstner} a scalar value is encoded in a spike train in terms of the frequency of spikes. Despite its effectiveness, this model presents some practical difficulties in explaining fast inference processes in the brain \citep{gerstner1996, thorpe2004}. In fact, in order to have reliable frequency coding, a large number of spikes need to be processed, which requires a long time.
Temporal coding \citep{gerstner1996} addresses this problem by encoding information in the actual timing of spikes w.r.t. some reference. In particular, Time-To-First-Spike (TTFS) coding uses the first spike only to encode information with minimal delay \citep{park2020, goltz2021, zhou2021}. The idea is that the more a neuron is stimulated, the earlier it will spike. Inter-Spike Intervals (ISI) \citep{kreuz2007} encode a vector of values by relying instead on multiple spikes, where information is encoded in the delay between consecutive spikes.
Rank order coding \citep{thorpe1998, thorpe2004} is another strategy for fast processing that relies on a single spike per neuron. Given a layer of many neurons, the idea is that the relevant information can be encoded not in the absolute timing of spikes, but simply in the order in which they are generated from the various neurons.
Note that Traditional ANN can be considered as rate-coded models, where the output of a traditional artificial neuron corresponds to the output frequency of a spiking neuron. However, Spike coding principles based on time, rank, or frequency, only allow encoding positive values. In fact, biological neurons are differentiated into excitatory and inhibitory units, so that specialized units are responsible to represent values associated with different signs \citep{kubota2016}. 

\subsection{Biological and Neuromorphic Computing}  \label{sec:spiking:neuromorphic}

Thanks to the spike-based communication paradigm, biological neurons are extremely efficient in terms of energy requirements \citep{javed2010}. Energy efficiency is an important issue in modern deep learning \citep{badar2021}; hence, research is oriented toward different computing paradigms to support neural computation.

\begin{figure}
    \centering
    \includegraphics[width=0.3\textwidth]{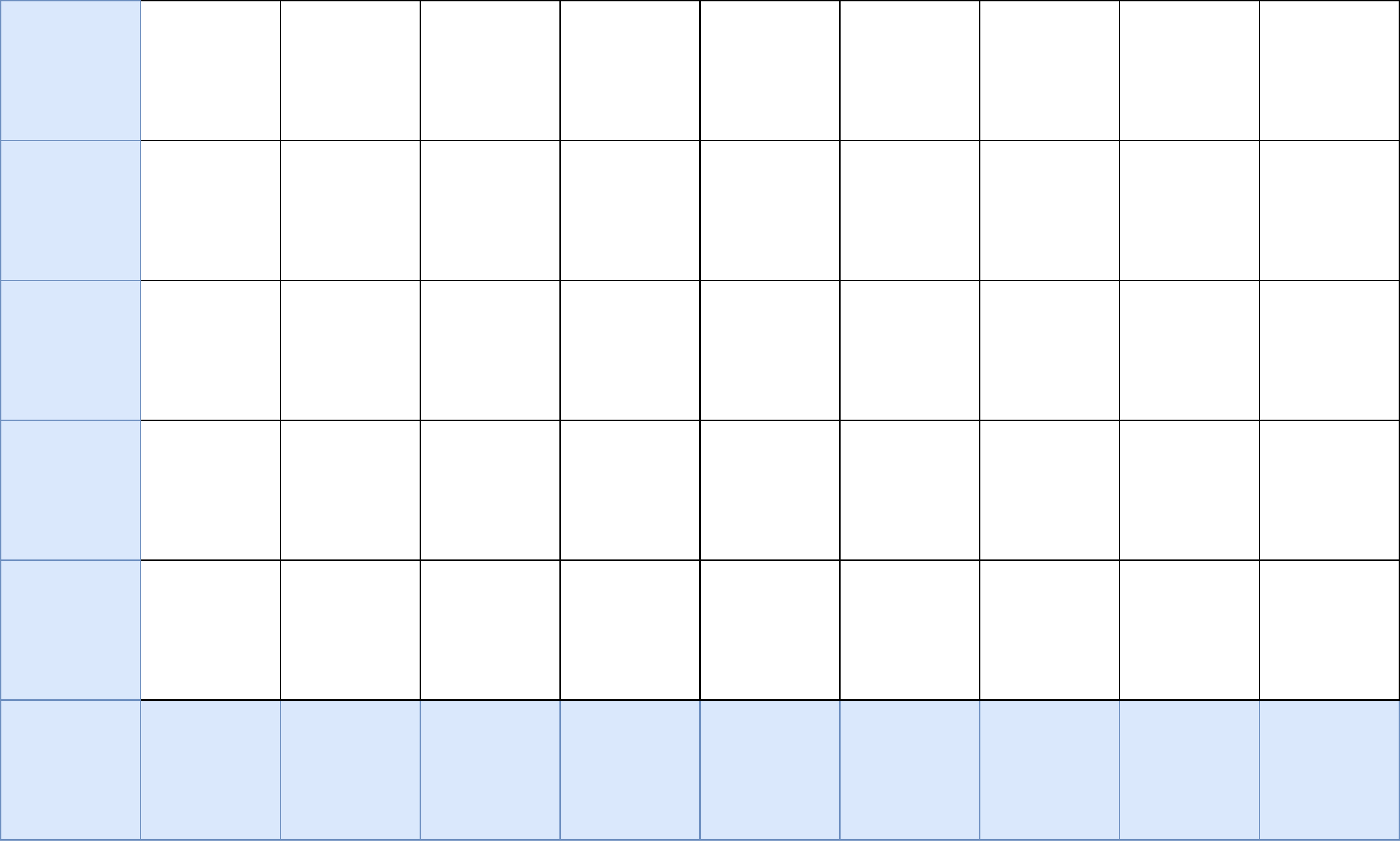}
    \caption{Example of electrode activations in a Multi-Electrode Array (MEA) grid forming an L-shaped pattern.}
    \label{fig:mea}
\end{figure}

Biological neural cultures could be leveraged as biological hardware (or \textit{bioware}) to address AI tasks \citep{ruaro2005, lagani2021a}. This allows to exploit the intrinsic spiking computation and plasticity mechanisms of real neurons, while promoting energy efficiency. There are various possible strategies to interface with biological neural cultures. For example, Multi-Electrode Arrays (MEA) \citep{gross1977, pine1980, bonifazi2002} are tiny grids of electrodes that can be used to stimulate or record neural activity at desired locations and time instants. Another possibility is represented by optogenetic channels, which enable interfacing with the neuronal culture through light-based stimulation \citep{meloni2020}. In order to harness the computational capabilities of real neurons, the main problem is related to defining adequate training protocols to induce plasticity in neurons, in order to instill the desired input-output relationship in the cultured network. A first approach in this direction was presented by Ruaro et al. \citep{ruaro2005}, where two types of stimuli were delivered to the networks through the electrodes of a 6x10 MEA grid: an L-shaped pattern \ref{fig:mea}, or its transpose. Stimuli were delivered by means of pulses on the desired electrodes at specified times. Initially, no significant difference in the network response to the two patterns (measured by counting the number of spikes generated by the neurons in response to each pattern) is observed. Then, a training phase takes place, in which the L-shaped pattern is repeatedly presented with high-frequency stimulation (\textit{tetanization}). After tetanization, the two patterns are presented again to the network, and it was observed that thanks to Hebbian plasticity, the network developed a stronger response to the L-shaped pattern. Therefore, it was shown that the network could develop pattern recognition capabilities. Recent work suggests a protocol to train biological networks also for more complex pattern recognition tasks \citep{lagani2021a}. The authors developed a biologically realistic simulator, based on SNN models, to create a digital twin of the neuronal culture, and it was shown that simulated cultures could achieve promising results in digit recognition tasks. Software simulation represents a valid tool to guide the design process of real-world cultures, allowing to find optimal culture parameters for a given task, that can later be reproduced \textit{in vitro}.
Biological cultures interfaced with MEAs were also used in another work \citep{kagan2022}, where they were trained to play simple games such as Pong.

Spiking neuron models represent a promising computing paradigm due to the possible applications in the implementation of computing hardware that reproduces the behavior of biological neurons, not only \textit{in vitro}, as in the case of bioware, but also \textit{in silico}, with devices known as \textit{neuromorphic hardware} \citep{roy2019b, zhu2020, schuman2022, huynh2022, shrestha2022}. By reproducing the spike-based computation in hardware, researchers were able to develop extremely energy-efficient neuromorphic chips \citep{gamrat2015, wu2015}, such as Neurogrid \citep{benjamin2014}, TrueNorth \citep{merolla2014}, BrainScales \citep{schemmel2010, billaudelle2020}, Loihi \citep{davies2018}.
Despite the energy-efficient computing paradigm, SNN models have to face novel challenges, compared to traditional DNNs, related to the learning and optimization paradigms. In fact, traditional learning based on backprop is not adequate for SNN, because the spiking nonlinearity is not well suited for gradient-based optimization. Therefore, the following subsections are dedicated to addressing the learning problem in SNNs.

\subsection{Plasticity in SNNs}  \label{sec:spiking:plasticity}

\begin{figure}
    \centering
    \begin{tikzpicture}
        \begin{axis}[
            width=0.3\textwidth,
            axis x line=middle,
            axis y line=middle,
            xlabel = {$\Delta t$},
            ylabel = {$\Delta w$},
            x tick label style={major tick length=0pt},
            y tick label style={major tick length=0pt},
            xticklabels={,,},
            yticklabels={,,},
        ]
            \addplot[
                domain=0:3,
                samples=30,
            ]{exp(-x)};
            \addplot[
                domain=-3:0,
                samples=30,
            ]{-exp(x)};
        \end{axis}
    \end{tikzpicture}
    \caption{Profile of the STDP weight update law.}
    \label{fig:stdp_update}
\end{figure}
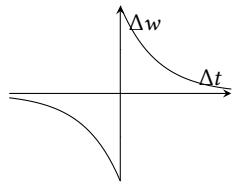

In biological spiking neurons, the equivalent of the Hebbian learning principle is represented by the \textit{Spike Time Dependent Plasticity} (STDP) rule \citep{bi1998, song2000}. In STDP, synaptic potentiation occurs when an input spike is received on a synapse and it is immediately followed by an output spike (while synaptic weakening is associated with spikes in reverse order). Specifically, a possible STDP rule can be expressed as follows:
\begin{equation} \label{eq:stdp}
\Delta w = \begin{cases}
    \eta^+ \, e^{-|\Delta t|/\tau^+} & \text{if $\Delta t > 0$} \\
    \eta^- \, e^{-|\Delta t|/\tau^-} & \text{otherwise}
\end{cases}
\end{equation}
where $w$ is the weight, $\Delta t$ is the time difference between the post-synaptic and the pre-synaptic spike, $\eta^+$ and $\eta^-$ are learning rate parameters ($\eta^+ > 0$ and $\eta^- < 0$) and $\tau^+$ and $\tau^-$ are time constants. Weight strengthening occurs when a pre-synaptic spike has been a likely cause for a post-synaptic spike, hence pre-synaptic and post-synaptic activations are correlated. If, instead, a pre-synaptic spike occurred right after a post-synaptic one, then the two activations are anti-correlated and a weight decrease occurs. Figure \ref{fig:stdp_update} shows the profile of a possible function that determines the dependency between $\Delta w$ and $\Delta t$ according to the STDP weight update rule. In this perspective, STDP realizes the Hebbian principle in the context of spiking neurons.

STDP can be expressed in terms of a \textit{mechanistic model} \citep{pfister2006}, where it is assumed that synaptic plasticity is realized in neurons by means of chemical agents that are released upon spiking events at the inputs and output of a neuron. No specific assumption about the nature of these chemicals is made, but their role is to act as firing event detectors. Let's assume a chemical agent, with concentration $c_x(t)$ over time, is released upon a firing event at an input synapse, while a chemical with concentration $c_y(t)$ is released upon an output firing event. The concentrations of the two chemicals vary in time as follows:
\begin{equation}
\begin{split}
    \frac{d c_x(t)}{d t} & = -\frac{1}{\tau_x} c_x(t) + \sum_{t_{in}} \delta(t - t_{in}) \\
    \frac{d c_y(t)}{d t} & = -\frac{1}{\tau_y} c_y(t) + \sum_{t_{out}} \delta(t - t_{out})
\end{split}
\end{equation}
where the terms in the sums represent the input and output spike trains, while $\tau_x$ and $\tau_y$ are time constants for the exponential decay of chemical concentrations. Plasticity occurs upon input or output firing events, depending on the concentrations of the output and input chemicals, according to the following rule:
\begin{equation} \label{eq:stdp_mechanistic}
    \Delta w = 
    \begin{cases}
        \eta_+ c_x(t) & \text{upon postsynaptic spike}\\
        \eta_- c_y(t) & \text{upon presynaptic spike}
    \end{cases}
\end{equation}
Upon postsynaptic spike, LTP is triggered proportionally to the concentration of $x(t)$, which is high if presynaptic events have occurred recently. Upon presynaptic spike, LTD is triggered proportionally to the concentration of $y(t)$, which is high if postsynaptic events have occurred recently. Hence, pre-before-post spikes are implementing LTP, while post-before-pre spikes are implementing LTD, according to STDP. Specifically, the chemical agents are implementing the exponential learning window of STDP, given their exponential decay. The mechanistic model allows us to establish a close relationship between STDP and rate-based Hebbian models. When the firing intervals are short enough w.r.t. the chemical time constants (high firing rate assumption), chemical concentrations are performing a running average of the input and output spike trains. Thus, the chemicals are basically tracking the input and output firing rates and encoding them in their concentrations (possibly up to a proportionality constant). In this scenario, LTP triggered upon output spikes is proportional to the input firing rate; but output spike events occur at the output firing frequency, so the overall LTP will be proportional to both the input and the output firing rate, thus implementing rate-based Hebbian learning. The same reasoning applies to LTD, in which case anti-Hebbian plasticity is implemented. The amplitudes of the learning rates $\eta_+$ and $\eta_-$ can be tuned in order to make the Hebbian behavior prevail over the anti-Hebbian behavior or vice-versa. This kind of mechanistic model has been used to define STDP rules based on more than two spikes, e.g. using triplets of spikes, by considering additional chemicals in the model \citep{pfister2006}. In general, using this approach it is possible to create learning terms in the spiking model that can be mapped to corresponding learning terms in the Hebbian rate-based model, e.g. terms proportional to $x$, $y$, $xy$, $x^2$, $y^2$, $xy^2$, $x^2 y$ as in Eq. \ref{eq:hebbian_general}.

Another class of models describes plasticity in SNNs not only in terms of spike times but also based on the value of the neuron membrane potential at a given instant. These models are referred to as Voltage-Dependent Synaptic Plasticity (VDSP), or Spike-Driven Synaptic Plasticity (SDSP) \citep{fusi2000, garg2022}. For example, in \citep{fusi2000}, SDSP is described as:
\begin{equation}
    \Delta w(t) = \sum_{t_pre} \phi(V(t_{pre})) \delta(t - t_{pre}) 
\end{equation}
where $V(t)$ is the value of the membrane potential and time $t$ and $\phi()$ is a nonlinearity analogous to that of the BCM rule, such that LTP occurs when $V(t)$ is sufficiently high. This behavior is related to STDP in the sense that a high value of the membrane potential at a given instant of time is related to a higher probability of firing soon in the future, while a low membrane potential denotes that the neuron has probably fired recently, but this rule does not require to maintain information about the timing of firing events or indicators such as chemical concentrations. 

Other variants of STDP rules have been proposed in the literature. 
Masquelier and Thorpe \citep{masquelier2007} propose a simplified STDP model where potentiation and depression only depend on the order of input and output spikes, but not on the relative timings:
\begin{equation} \label{eq:bistable_stdp}
\Delta w = 
    \begin{cases}
        \eta_+ w (1 - w) & \text{LTP}\\
        \eta_- w (1 - w) & \text{LTD}
    \end{cases}
\end{equation}
In order to prevent the weights from becoming too large or too low, soft-thresholding mechanisms are introduced by modulating the updates by $w$ and $1-w$ terms.
An alternative form is presented in \citep{nessler2009} (and further explored in \citep{tavanaei2016}) which is related to the Expectation-Maximization algorithm. In this case, neuron models with noisy thresholds are used, and the corresponding updates are in the form:
\begin{equation} \label{eq:em_stdp}
\Delta w = 
    \begin{cases}
        \eta_+ e^{-w} & \text{LTP}\\
        \eta_- e^{ w} & \text{LTD}
    \end{cases}
\end{equation}
In \citep{wade2010}, Synaptic Weight Association Training (SWAT) is proposed as a spiking variant of the BCM rule used in conjunction with STDP. The BCM-like behavior is achieved by an STDP rule \ref{eq:stdp} with learning rate parameters $\eta_+$ and $\eta_-$ dynamically chosen as a function of a threshold parameter $\theta$:
\begin{equation}
\begin{split}
    \eta_+ = \eta_0 \frac{1}{1 + \theta}
    \eta_- = \eta_0 - \eta_+
\end{split}
\end{equation}
where $\eta_0$ is a base learning rate parameter and $\theta = (\frac{y}{c_0})^\alpha y$.
Here, $y$ is the output firing rate and $\alpha$ (=2) and $c_0$ are constants. 
The authors also propose a modified rule in which a threshold is associated with each synapse, instead of computing an overall threshold based on the neuron output. This new threshold is chosen as a function of the current weight value on the synapse $\theta = (\frac{w}{c_0})^\alpha w$. The thresholding behavior is intended to adapt the learning rate in order to favor learning when the neural activity is sufficiently high.

\subsection{STDP for Deep Learning}  \label{sec:spiking:dl}

Supervised training has been an essential tool in the development of state-of-the-art deep learning solutions \citep{he2016, devlin2019, silver2016}. However, supervised training in deep architectures requires the backpropagation mechanism to assign errors to single computational units. However, in SNNs the backprop mechanism is implausible, due to the nature of the spiking nonlinearity, which has zero derivatives almost everywhere. 
Therefore, STDP-based local learning rules have been used as an alternative to train networks for DL tasks, such as image recognition.

In a pioneering work \citep{diehl2015b}, STDP learning was used to train a network with a single convolutional layer of WTA competing units for digit recognition tasks, and an experimental evaluation was performed. WTA competition in spiking neurons can be implemented by lateral inhibition, such that the first spiking neuron inhibits the others. Classification based on the spiking representations extracted by the trained layer achieved 95\% accuracy on MNIST \citep{mnist}.

\begin{figure}
    \centering
    \includegraphics[width=0.2\textwidth]{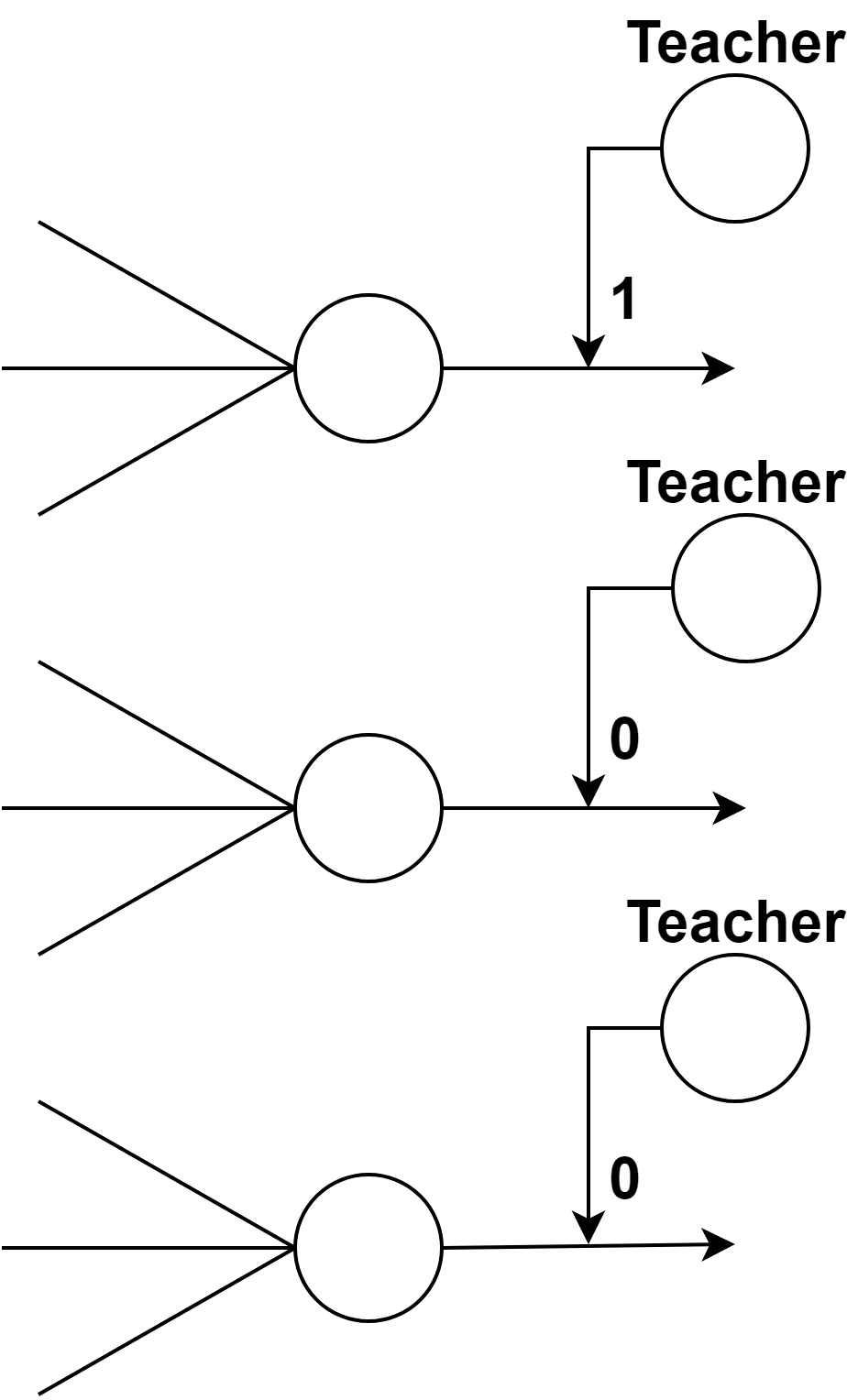}
    \caption{Teacher neurons inducing a desired output on real neurons.}
    \label{fig:teacher}
\end{figure}

Although STDP is an unsupervised algorithm, supervision can be achieved by means of the \textit{teacher neuron} technique \citep{ponulak2005}. In \citep{shrestha2017b}, a variant of the teacher neuron technique was used, which consists in imposing a teacher signal on a neuron and performing STDP updates (Fig. \ref{fig:teacher}). In this way, if the neuron is strongly excited by the teacher's signal, it is induced to learn to respond strongly to patterns similar to the current input; otherwise, learning is inhibited. The authors trained a network with a single convolutional hidden layer, followed by a classification layer. Teacher neurons were used to train the output layer, while the hidden layer was trained using soft-WTA and a simplified version of the STDP rule called Quantized 2-Power Shift (Q2PS), expressed by the following update equation:
\begin{equation}
\Delta w = 
    \begin{cases}
        \eta_+ 2^{-w} & \text{LTP}\\
        \eta_- 2^{ w} & \text{LTD}
    \end{cases}
\end{equation}
This is a variant of Eq. \ref{eq:em_stdp}. The difference is that exponentials are replaced by powers of 2, so that the rule can be implemented more efficiently by means of shift operations. Experimental results showed and accuracy of 89.7\% on MNIST digit recognition. The teacher-neuron approach was also adapted to the case of Hebbian neural networks \citep{lagani2019, mthesis, lagani2021a}.

In \citep{kheradpisheh2018}, a network composed of three convolutional layers trained with the variant of STDP in Eq. \ref{eq:bistable_stdp} \citep{masquelier2007}, and preceded by a layer of DoG (Difference of Gaussian) filtering, was used. The STDP variant enforced the weights to remain between 0 and 1 (\textit{soft thresholding}). The extracted features were fed to an SVM classifier. The model achieved 82.8\% accuracy on the ETH-80 \citep{eth80} dataset and 98.4\% accuracy on MNIST.

In \citep{ferre2018}, a hybrid spiking-traditional architecture was proposed. WTA was used in conjunction with STDP to train a network composed of two hidden convolutional layers. The features extracted from these layers were then fed to another traditional ANN branch composed of other two fully connected hidden layers, followed by a final classification layer, trained with backpropagation. The resulting architecture achieved 98.49\% accuracy on MNIST. The same approach was used to solve the ETH-80 image classification task, this time using a network with 3 convolutional layers trained with STDP followed by a single fully-connected layer trained with backprop, achieving 75.2\% accuracy. Experiments were also performed on CIFAR-10 \citep{cifar} and STL-10 \citep{stl10} datasets, using an architecture with a single STDP convolutional layer, two fully connected backprop layers, and the final classification layer, also trained with backprop. The method achieved 71.2\% accuracy on CIFAR-10 and 60.1\% accuracy on STL-10.

In addition to synaptic weight modification, biological neurons also exhibit other forms of adaptation that are collectively referred to as \textit{homeostatic} plasticity \citep{watt2010, turrigiano2012}. For example, the membrane potential time constant and firing threshold are among the parameters of spiking neurons that can be subject to adaptation. Threshold adaptation in SNN was introduced in \citep{falez2019}, as a means to normalize the spiking activity to a target average firing rate. The authors performed experiments with a network of 3 convolutional layers, with an accuracy of 98.6\% on MNIST.
Ensembling of multiple SNN models can also be beneficial, as in \citep{fu2021}, where an ensemble of 9 SNNs was able to achieve MNIST accuracy of 98.7\%. Model compression techniques for SNNs have also been explored, such as in \citep{putra2022}, where an SNN model with a single excitatory and a single inhibitory layer was explored, and model compression to less than 200MB space was still able to achieve 92\% accuracy on MNIST.

\begin{table}
    \centering
    \caption{Experimental results of bio-inspired STDP methods for deep learning applications on the MNIST dataset.}
    \label{tab:stdp_dl}
    \begin{tabular}{c|p{0.5\textwidth}|c}
         \textbf{Method} & \textbf{Description} & \textbf{MNIST Acc. (\%)} \\
         \hline
         STDP + WTA & Single conv layer trained by competitive learning and STDP, followed by an SVM classifier. & 95\% \citep{diehl2015b} \\
         Q2PS + WTA + teacher neuron & Fully-spiking architecture: 1 conv layer followed by a teacher neuron classifier. & 89.7\% \citep{shrestha2017b} \\
         STDP-CNN & Experimental analysis of STDP variant from \citep{masquelier2007} on a 3-layer CNN followed by an SVM classifier. & 98.4\% \citep{kheradpisheh2018} \\
         Hybrid STDP-backprop & Hybrid combination of spiking and traditional layers. A sequence of spiking convolutional layers trained by STDP + WTA is used to extract features that are then fed to a traditional network branch trained by supervised backprop. 2 spiking conv layers followed by 2 traditional fc layers. & 98.48\% \citep{ferre2018} \\
         STDP with threshold adaptation & 3 conv layers with homeostatic adaptation of membrane potential firing threshold & 98.6\% \citep{falez2019} \\
         SNN ensemble& Ensemble of SNNs trained with STDP & 98.7\%  \citep{fu2021} \\
         TinySNN & SNN with a single excitatory and a single inhibitory layers, compressed to < 200MB & 92\% \citep{putra2022} \\
    \end{tabular}
\end{table}

The results discussed above are summarized in Tab. \ref{tab:stdp_dl}

\section{Backpropagation Alternatives} \label{sec:bp_alternatives}

This Section is dedicated to the discussion of alternative algorithms for supervised neural network training, without leveraging the traditional backprop approach. Following the discussion from the previous Section, we start by considering the challenges of backprop-based optimization in SNNs. We then follow with a discussion of backprop-free approaches that rely on local Hebbian-type updates and reward modulation. Another class of methods also uses Hebbian updates, together with top-down connectivity patterns. Finally, while previous methods focus on approximating backprop, while facing structural or computational constraints, another class of methods will be discussed, targeting the learning problem from a completely different perspective.

In the following, subsection \ref{sec:bp_alternatives:snn} discusses the challenges and solutions of adapting gradient-based optimization to SNNs; subsection \ref{sec:bp_alternatives:reward} illustrates learning methods based on reward assignment and Hebbian plasticity which can be shown to approximate backprop; alternatively, backprop approximation by means of Hebbian updates can also be achieved in networks with top-down feedback connectivity, which is the topic of subsection \ref{sec:bp_alternatives:feedback}; finally, subsection \ref{sec:bp_alternatives:beyond} presents alternative approaches that tackle the learning problem from different perspectives, without resorting to backprop approximations.

\subsection{Backprop Approaches for SNN Training} \label{sec:bp_alternatives:snn}

In SNNs, the backprop mechanism is implausible due to the nature of the spiking nonlinearity, which has zero derivatives almost everywhere. In order to bring the power of gradient-based optimization to SNNs, researchers have proposed methods to convert traditional pre-trained ANNs to SNNs \citep{diehl2015a, rueckauer2017, hu2018, sengupta2019}. Direct training methods, on the other hand, \citep{neftci2019, lee2016, wu2018, wu2019, lee2020} try to circumvent the differentiability problems with surrogate gradients, which leverage approximations of the backward pass through the spike generation process. This is done by smoothing the spike nonlinearities, typically with exponentially decaying functions, so that the smoothed function has useful derivatives. The resulting methods are analogous to an STDP rule with error modulation \citep{tavanaei2019b, shrestha2019}. 

The first work in which the idea of backpropagation learning was applied to SNNs is SpikeProp \citep{bohte2000}. The main issue to solve in this case is how to compute derivatives of spiking signals, given that pulses are not differentiable. The authors proceeded as follows: first, given the desired output spike times $ t_J^d $, an error function is defined:
\begin{equation}
    E = \frac{1}{2}\sum_j(t_j - t_j^d)^2
\end{equation}
Then, the derivative of $E$ w.r.t. weight $w_{ij}$ is
\begin{equation}
    \frac{\partial E}{\partial w_{ij}} = \frac{\partial E}{\partial t_j} \frac{\partial t_j}{\partial x_j(t)} \Big{|}_{t=t_j} \frac{\partial x_j(t)}{\partial w_{ij}} \Big{|}_{t=t_j}
\end{equation}
Where the output firing time $t_j$ has been expressed as function of the total postsynaptic input $x_j$. Similarly, the derivative of E w.r.t. a presynaptic input, which corresponds to the error backpropagated to the previous layer, is
\begin{equation} \begin{split}
    \frac{\partial E}{\partial x_i(t)} & = \sum_j \frac{\partial E}{\partial t_j} \frac{\partial t_j}{\partial x_j(t)} \Big{|}_{t=t_j} \frac{\partial x_j(t)}{\partial x_i(t)} \Big{|}_{t=t_j} \\
    & = \sum_j \frac{\partial E}{\partial t_j} \frac{\partial t_j}{\partial x_j(t)} \Big{|}_{t=t_j} w_{i, j}
\end{split} \end{equation}
At this point the assumption is made that, for a small region $\epsilon$ around $t_j$, the dependence of $t_j$ w.r.t. $x_j$ is approximately linear. Thus, we can write
\begin{equation}
    \delta t_j \simeq -\delta x_j(t_j)/\alpha
\end{equation}
where $\alpha = \frac{\partial{x_j(t)}}{\partial{t}}\Big{|}_{t=t_j}$. Thus,
\begin{equation}
    \frac{\partial t}{\partial x_j(t)} \Big{|}_{t=t_j} = -\frac{1}{\alpha} = \frac{-1}{\frac{\partial{x_j(t)}}{\partial{t}}\Big{|}_{t=t_j}}
\end{equation}
and we have all the ingredients to perform gradient descent. 
Improvements to this approach for handling multiple output spikes have also been proposed, such as MultiSpikeProp \citep{booij2005}. Some approaches, such as the Chronotron \citep{florian2012}, rely on special spike error metrics (such as the Victor-Purpura metric \citep{victor1997}), while others aim at matching the output spike time probability distribution with a desired distribution \citep{brea2011, rezende2011}. 
 
Note that in the rule above, the gradient depends on the value of the parameter $\alpha$, which is in turn related to the behavior of the membrane potential at spike arrival times. If we simply sum up the spikes as soon as they arrive, the membrane potential becomes non-differentiable in time, which corresponds to infinite $\alpha$. One way to cope with the non-differentiability is to use a smoothing kernel for the spike nonlinearities, as it is also done in the SPAN, PSD, ReSuMe and SuperSpike algorithms \citep{mohemmed2012, yu2013, ponulak2005, zenke2018}. 
In \citep{liu2017a}, a different point of view was taken, in which synaptic parameters were considered as tunable delay elements. It was possible to formulate a gradient descent update on these parameters in order to allow the networks to reproduce the desired output spike time patterns.

The \textit{teacher neuron} technique, already presented in the previous Section, was first proposed in ReSuMe \citep{ponulak2005} as a method to tune spiking networks to generate desired spike patterns. The original teacher neuron techniques requires performing LTP when a teacher spike occurs, and LTD after an ordinary output spike. In other strategies, the teacher neuron process is sometimes simplified by removing the LTD part, while the role of synaptic depotentiation is often taken by a weight decay term, such as that arising in WTA methods \citep{shrestha2017b, lagani2019, lagani2022a}. 

An issue of the approaches described above is that we have to provide the precise target output firing times, rather than, for example, simple class labels. Therefore, several variations and improvements have been proposed. In some applications, the requirement to generate exact spike patterns might be too strict. For example, we might be interested to count the number of spikes generated by the neurons, in response to a given input, while ignoring their precise timing. In this case, an appropriate error measure could be the difference between desired and actual spike counts at every instant ($d_j(t) - y_j(t)$), rather than the difference in spike times ($t_j^d - t_j$). Using ($d_j(t) - y_j(t)$) as a factor to modulate STDP updates allows minimizing the error given by the squared difference between desired and actual output spike count at every time instant. This kind of modulated STDP emerges in some of the works mentioned above \citep{florian2012, mohemmed2012, yu2013, zenke2018}, as well as in another interesting approach, called BP-STDP \citep{tavanaei2019b}. A closer look at the latter approach can help us to get insights on the quantity $d_j(t) - y_j(t)$. In BP-STDP, time is divided into little slots in which a spike at a given synapse or axon either can or cannot occur. In every time slot, the error is defined to be proportional to $(d_j(t) - y_j(t))^2$, which is 1 when an output spike is desired, but none is produced, it is -1 when an output spike is produced, but none is desired, or it is 0 otherwise. The error can be minimized by the following learning rule:
\begin{equation}
    \Delta w_{i, j}(t) = \eta (d_j(t) - y_j(t)) \cdot spike \; count
\end{equation}
where $spike count$ is the number of input spikes a given synapse in a time slot. This rule performs STDP-like updates when an input spike and a desired output spike occur simultaneously, and anti-STDP-like updates when an input spike and an actual output spike occur simultaneously. Alternatively, the rule can be interpreted as a modulated STDP rule in which updates take place when input and output spikes occur simultaneously, and the modulation factor $(d_j(t) - y_j(t))$ is 1 when an output spike is desired but not actually produced, -1 when no output spike is desired but one is actually produced, 0 otherwise. 

Approaches based on spike-count, rather than precise firing times, are more flexible and better suited for some types of problems, for example in classification problems, when a category label is available as a teacher signal. In a more recent work \citep{lee2016}, further extended in \citep{lee2019}, training based on categorical labels is achieved as follows: output spikes are integrated over a time interval T and then divided by the number of time step in T, in order to obtain a numerical value for the output which is then used for classification. 
\begin{equation}
    output = \frac{num \; spikes (T)}{num \; time \; steps}
\end{equation}
This is equivalent to computing errors w.r.t. rate coded outputs. In this case, a simple approximation is made for the parameter $\alpha$, which is taken to be equal to the reciprocal of the firing threshold:
\begin{equation} 
     \frac{\partial t_j}{\partial x_j}\Big{|}_{t=t_j} = \frac{1}{V_{th}} 
\end{equation}
Other approximations are also made to account for the leaky behavior of a LIF neuron. Networks trained with this approach reach accuracy comparable with traditional DNN models, but suitable for implementation in energy-efficient neuromorphic computing. Results close to the latter are achieved in other works in which similar approaches are taken \citep{wu2018, wu2019, zheng2020, rathi2021, wu2021a, wu2021b}, where similar rate coding schemes are used for the output, but the spiking differentiation problem is solved by using smoothing kernels for the spikes. Specific approaches address the problem of achieving fast inference, in addition to accurate predictions, which can be obtained by leveraging TTFS coding, as in \citep{park2020, zhou2021}. In these methods, spiking-backprop learning approaches are adapted to induce the first spike timing generated by the network to match a precise desired firing time. A hybrid approach is taken in \citep{rathi2020}, where a fine-tuning based on a spike-based backpropagation approach \citep{wu2018} is performed after the conversion of a pre-trained DNN to SNN. 

\begin{table}
    \centering
    \caption{Experimental results of alternative methods for SNN training on the CIFAR-10 dataset.}
    \label{tab:snn_train}
    \begin{tabular}{p{0.3\textwidth}|p{0.5\textwidth}|c}
        \textbf{Method} & \textbf{Description} & \textbf{CIFAR-10 Acc. (\%)} \\
        \hline
        Spatio-Temporal BackProp (STBP) & 7-layer AlexNet-like CNN (CIFARNet) direct training. & 90.53\% \citep{wu2018, wu2019} \\
        Spiking backprop & ResNet11 direct training. & 90.95\% \citep{lee2016, lee2020} \\
        STBP with threshold-dependent normalization (tdSTBP) & ResNet19 direct training. & 93.10\% \citep{zheng2020} \\
        Accumulated spiking flow & VGG7 direct training. & 91.35\% \citep{wu2021a} \\
        Tandem learning rule & AlexNet-like 7-layer CNN trained with an auxiliary coupled DNN with a nonlinearity derived to match the rate-coded spiking nonlinearity for gradient backprop. & 91.77\% \citep{wu2021b} \\
        T2FSNN & TTFS coding with temporal backpropagation for precise firing time tuning on a VGG16 architecture. & 91.43\% \citep{park2020} \\
        Temporal coded deep SNN & TTFS coding where a closed-form expression of firing time is derived and used for optimization on a VGG16 architecture. & 92.68\% \citep{zhou2021} \\
        DIET-SNN & VGG16 direct training with optimization of leakage and threshold parameters. & 93.44\% \citep{rathi2021} \\
        DNN-SNN conversions & Conversion of a 6-layer AlexNet-like CNN. & 88.82\% \citep{rueckauer2017} \\
        Conversion of residual models & DNN-SNN conversion of ResNet44. & 92.37\% \citep{hu2018} \\
        Conversion of deep models & DNN-SNN conversion of VGG16. & 91.50\% \citep{sengupta2019} \\
        Hybrid conversion + STBP & DNN-SNN conversion + STBP fine-tuning on VGG16. & 92.02\% \citep{rathi2020} \\
    \end{tabular}
\end{table}

Table \ref{tab:snn_train} summarizes the methods presented so far for SNN training, together with some experimental results on the CIFAR-10 dataset.

\subsection{Models Based on Reward Modulation} \label{sec:bp_alternatives:reward}

A biologically plausible alternative to error-driven learning is reward-driven learning. In \citep{williams1992} the REINFORCE algorithm was presented. This is a pioneering algorithm for reward-driven learning in networks of stochastic units. In REINFORCE, updates are performed according to the following rule:
\begin{equation}
    \Delta w = \eta (r - b) e
\end{equation}
where r is the reward signal, b is a baseline parameter and e is called the \textit{eligibility trace} of $w$: $ e = \partial ln(g(y | net)) / \partial w$. The function $g$ is the probability density function of the stochastic neuron to emit output $y$ given its current input $net$. It is shown that the algorithm produces updates, on average, in the direction that maximizes the reward, as gradient ascent would do. A typical problem in real environments is that the reward might come sometime after a decision was made (distal reward problem). The authors propose a solution by using a running sum of the eligibility traces over time. REINFORCE updates are essentially of Hebbian type, but enhanced by reward modulation. Generalizations of this approach form the category of three-factor Hebbian learning models \citep{gerstner2018}

Since the REINFORCE provides update directions that estimate the gradient direction with high variance, the convergence speed scales poorly with the number of units in the network. Possible improvements have been proposed, based on gating mechanisms aimed at identifying specific neurons to update \citep{roelfsema2005, pozzi2018}. In \citep{roelfsema2005}, the Attention-Gated REinforcement Learning (AGREL) algorithm was proposed as an extension of REINFORCE. AGREL also follows reward-driven updates, but it adds feedback signals (attention gating mechanism) to determine which neuron in a given layer contributed the most to activate neurons in the next layer. This is done by choosing the neuron that provided the highest output in the last layer as winner and setting its gating factor to 1, while the gating factors of the other neurons are set to 0; then, the attention gating signal propagated backward using the transpose of the weight matrix, as backpropagation would do. The resulting algorithm is shown to be related to backpropagation. A variant of this rule, named Q-AGREL, was used in \citep{pozzi2018} to address computer vision tasks. Kickback \citep{balduzzi2015} is another strategy based on reward modulation, which leverages particular constraints in the network organization, related in particular to the sign of connections from a neuron to the output, to make sure that a neuron is always updated in a direction that optimizes the objective.

A model extending reward-modulated learning in the context of SNNs was proposed in \citep{florian2005}. The authors derived the following learning rule for networks of stochastic spiking neurons (based on the noisy threshold model):
\begin{equation}
    \Delta w(t) = \eta r(t) z(t)
\end{equation}
Where $z(t)$ is the eligibility trace, computed online as:
\begin{equation}
    z(t+1) = \beta z(t) + \zeta(t)
\end{equation}
with 
\begin{equation}
    \zeta(t) = 
    \begin{cases}
        \frac{1}{\sigma(t)}\frac{\partial \sigma(t)}{\partial w} & \text{if neuron fires at time t} \\
        -\frac{1}{1-\sigma(t)}\frac{\partial \sigma(t)}{\partial w} & \text{otherwise}
    \end{cases}
\end{equation}
In this expression, $\sigma(t)$ is the probability of the stochastic spiking neuron to fire at time t.
Finally, an experimental evaluation of reward-modulated STDP (R-STDP) learning in computer vision tasks was provided in \citep{mozafari2018}.

\subsection{Top-Down Feedback Models} \label{sec:bp_alternatives:feedback}

One of the reasons for considering backprop biologically implausible is the \textit{weight transport problem}, i.e. in order to deliver the error signal to each neuron, the network should be endowed with symmetric backward connections \citep{richards2019}. Backward connections are present in the brain, but imposing the symmetry constraint is not biologically plausible. 

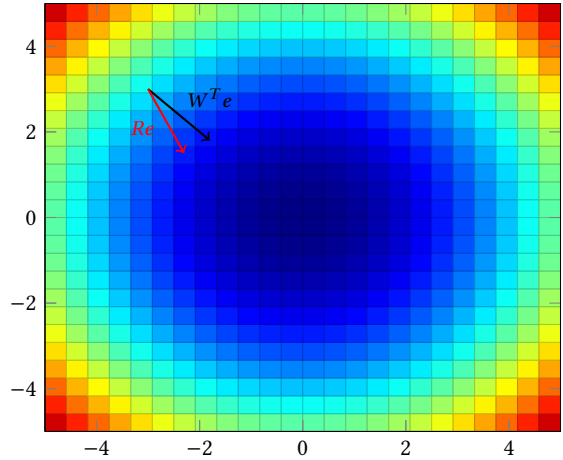
\begin{figure}
\centering
\begin{tikzpicture}
	\begin{axis}[view={0}{90}]
		\addplot3[  surf, 
		            colormap/jet
		        ] {x^2 + y^2};
		\draw[->, thick](axis cs: -3, 3) -- (axis cs: -1.8, 1.8);
	    \node[] at (axis cs: -1.8, 2.8) {\small{$W^Te$}};
	    \draw[->, thick, red](axis cs: -3, 3) -- (axis cs: -2.3, 1.5);
	    \node[] at (axis cs: -3.1, 2.1) {\textcolor{red}{\small{$Re$}}};
	\end{axis}
\end{tikzpicture}
\caption{Approximation of gradient descent by alternative directions. Directions that are less than 90° away from the (negative) gradient are still descent directions in the loss landscape.}
\label{fig:approx_descent}
\end{figure}

A proposal for solving the weight transport problem was presented in \citep{lillicrap2014}, in which the authors replace symmetric backward weights with random ones. They show that, even with this choice, under appropriate assumptions, the network can still solve simple tasks and, in particular, the forward weights converge to the \textit{pseudo-inverse} of the random feedback weights. This approach is called \textit{Random Feedback Alignment}. The idea is that calling $W$ the weight matrix at some layer and $e$ an error vector to be backpropagated, if we backpropagate the error using a random matrix $R$ instead of $W^T$, we obtain $Re$ as a backpropagated vector instead of $W^Te$, which will still allow descending the error surface (fig. \ref{fig:approx_descent}) as long as $Re$ and $W^Te$ are less than $90^{\circ}$ apart, i.e. $WR$ is positive semidefinite.

However, in another work \citep{zenke2018} it has been shown that RFA, while successful for solving simple tasks, is not adequate for more complex scenarios. Indeed, the pseudo-inverse of the random feedback weight matrix is still something random and a network with random feedforward weights, while being still able to solve some tasks (see, for example, extreme learning machines \citep{huang2006}), cannot leverage the full potential of the deep architecture. RFA is further improved in another work \citep{amit2019} by initializing forward and backward connections randomly and non-symmetrically, but then performing symmetric updates on both, thus achieving an effective approximation of backpropagation.

\begin{figure}
    \centering
    \includegraphics[width=0.6\textwidth]{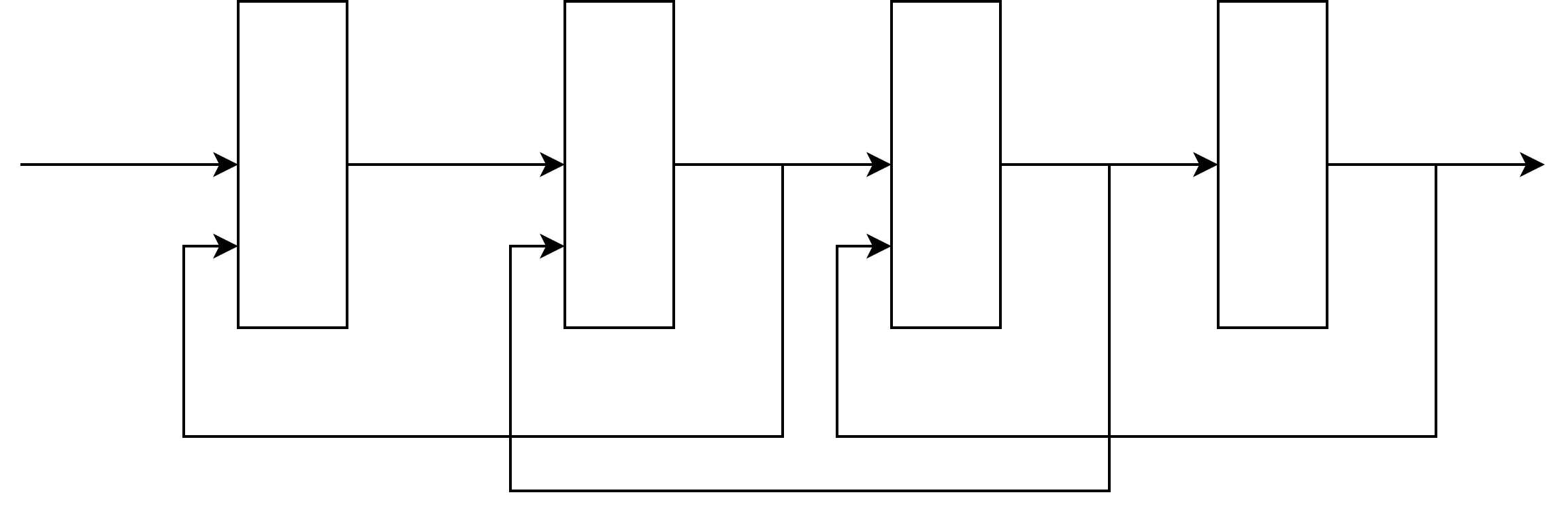}
    \caption{A network model with top-down connectivity}
    \label{fig:top_down_net}
\end{figure}

Even though RFA provides a mechanism to cope with the weight transport problem, this still requires a distinction between the types of connections for inference and error delivery, which is not supported in the brain \citep{oreilly}.
In order to address these problems, other models have been developed, which are able to approximate backprop with Hebbian updates, in neural networks with both feedforward and top-down connectivity (Fig. \ref{fig:top_down_net}), but without distinctions between the connections used for inference and error computation. As opposed to feedforward architecture, top-down connections form feedback loops that induce a recurrent computation in the network, whose dynamics can be modeled as an Energy-Based Model (EBM) \citep{haykin, hopfield1982, movellan1991, xie2003, scellier2017}. Although top-down connections are not typically used in DNNs, because unfolding the recurrent loops requires additional computing efforts, they are a pervasive feature in biological brains \citep{felleman1991, douglas1995}. Indeed energy-based recurrently connected architectures form the basis of biological associative memory models, such as the Hopfield network \citep{hopfield1982, wang2011b}. EBMs are dynamical systems whose state evolves by descending an energy landscape. These networks are able to store patterns in terms of minima in the energy landscape, so that when a perturbed version of a stored pattern is presented to the network, the dynamics will converge to a local minimum energy configuration, thus reconstructing the original pattern. It can be shown that Hebbian-like learning rules have the effect of reducing the energy in correspondence of a given input, thus providing a biologically grounded mechanism for the storage of memories and associations in EBMs \citep{hopfield1982, haykin}.

A promising approach in the category of EBM-based backprop alternatives is \textit{predictive coding}, which is also a popular neuroscientific theory \citep{harpur1996, rao1999}. Predictive coding networks \citep{wen2018, han2018} operate by having each layer trying to predict the output of the next layer, and, at the same time, approaching the prediction from the previous layer. This objective can be shown to result into EBM-type dynamic models of neural computation \citep{haykin}, and Hebbian models of synaptic modification which approximate backprop \citep{whittington2017, millidge2020}. In this context, the \textit{free energy} principle \citep{friston2005} is a neuroscientific theory, based on the EBM formulation of predictive coding, that describes both learning and inference from a common principle of minimization of a free energy function associated with the EBM. More biologically plausible models which also incorporate lateral inhibitory interaction have also been developed \citep{urbanczik2014, guerguiev2017, sacramento2018}. In predictive coding, every layer needs to compute the difference between its activation and the prediction from the previous layer, in order to determine both the activation and the synaptic dynamics. As in the Locally Competitive Algorithm (LCA) \citep{rozell2008}, lateral inhibition is essentially a more biologically realistic feature that replaces this difference computation. 
Backpropagating Action Potentials (BAP) \citep{williams2000} are further adopted in \citep{schiess2016} as a mechanism for backward error transmission through the neurons' dendritic trees.
In \citep{carreira2014, askari2018}, the backprop optimization objective is extended with auxiliary variables, which allows decoupling the optimization of each layer from the others, obtaining an optimization process equivalent to predictive coding.

Contrastive Hebbian Learning (CHL) approaches \citep{movellan1991, oreilly1996, xie2003, scellier2017} (not to be confused with contrastive \textit{representation} learning) train networks with top-down connectivity by alternating between two phases: a \textit{free phase}, and a \textit{clamped phase}. In the free phase, an input is provided to the network, and the recurrent dynamics in unfolded until convergence to a steady state. Let us denote with $x^-$ and $y^-$ the input and output activations of a given neuron during the free phase. In this phase, neurons perform an anti-Hebbian weight update $ \Delta w^- = -\eta y^- x^- $. After the free phase, the clamped phase takes place. During the clamped phase, while the same input is maintained, the output of the network is forced to the desired target value, represented by a teacher signal. Due to the recurrent connections, forcing the output values will affect the activations of all the neurons in the network. Again we wait for the network to reach a steady state. Call $x^+$ and $y^+$ input and output of a given neuron during the clamped phase. At this point, the network performs a Hebbian update $ \Delta w^+ = \eta y^+ x^+ $. It is possible to combine the updates of the two phases in a unique step, obtaining
\begin{equation}
    \Delta w = \eta (y^+ x^+ - y^- x^-)
\end{equation}
It can be shown that the resulting update approximates backprop under mild conditions \citep{movellan1991, oreilly1996}. The rationale behind the CHL rule is to update the network parameters in such a way that the next time the same input is provided, the network will tend to settle down into a steady state in which the network output corresponds to the desired output. For this purpose, the effect of an anti-Hebbian update is to increase the energy of the EBM in correspondence of the free-phase equilibrium, and the effect of the Hebbian update is to decrease the energy of the free-phase equilibrium so that the network will be more likely to settle down to the desired state without clamping.
It has been conjectured that the alternation between the two phases of CHL, if biologically implemented in the brain, could be the origin of Beta oscillations \citep{baldi1991}. In this context, The free phase would be a phase in which the brain computes the expected outcome of an action, while the clamped phase would be a phase in which the brain perceives the actual outcome of the action and uses the received signal as a teacher signal, thus providing also an explanation for the origin of supervision in the brain.

A recurrent network scheme has also been used in the Recirculation Algorithm \citep{hinton1988}, which can be used for autoencoder training. This algorithm was further extended to generic networks, leading to the Generalized Recirculation (GenRec) algorithm, which, under mild conditions, turns out to be equivalent to CHL \citep{oreilly1996}. The GenRec/CHL algorithm was further extended in \citep{oreilly2001} in order to improve the generalization performance. In Fact, it was noticed that networks with top-down feedback, although computationally more powerful than feedforward models, were prone to overfitting. The reason is that recurrent networks have chaotic attractor dynamics that, in some cases, might lead the system to settle down to a very different steady state when a slightly different input is provided. Another justification for the generalization difficulties in these models could be that recurrent network models corresponds to a deeper and more complex model w.r.t. purely feedforward counterparts, hence more prone to overfitting. The idea proposed in \citep{oreilly2001} for improving the GenRec generalization performance was to consider the advantages of both supervised and unsupervised learning. Supervised learning is useful to learn task-specific knowledge, while unsupervised learning is useful to learn generic knowledge about the data. Hence, error-driven supervised learning can help to solve a given task, while unsupervised learning can act as a regularizer. Hence, combining both supervised and unsupervised learning in the same rule could help us get the best of both worlds. Indeed, rather than a general principle, it is possible that multiple learning mechanisms \citep{burrone2002, morrison2008, clopath2010, zenke2015} are employed in the brain, to cope with diverse situations. The resulting algorithm, called LEABRA, first applies error-driven optimization according to GenRec, and then, while maintaining the activations of the clamped phase, a k-WTA unsupervised Hebbian update is applied. The authors show that the algorithm is actually useful for improving the generalization performance.

In \citep{xie2003}, a variation of the CHL approach was proposed, which used week feedback, so that the top-down computation only slightly modified (by a small factor $\epsilon$) the feedforward state, in order to make the recurrent model approximately equivalent to the simple feedforward model training. Similarly, in Equilibrium Propagation \citep{scellier2017}, only a weak clamping was provided, i.e. the natural output of the network was only slightly modified (again, by a small factor $\epsilon$) towards the desired target, thus realizing a more stable computation. 
While CHL looks more promising as a biologically plausible rule for learning w.r.t. gradient descent, it still requires the weights on the backward connections to be symmetric to the forward ones. Actually, this is not a problem in this case, because the CHL update rule is itself symmetric (i.e. it is identical if we exchange x and y), hence it spontaneously produces symmetric weights. But still, taking inspiration from RFA \citep{lillicrap2014}, it has been proven that CHL also can converge even when random feedback weights are used \citep{detorakis2019}.
Extensions to SNNs of top-down feedback-based backprop approximations have also been proposed \citep{neftci2014, oconnor2019}. In \citep{neftci2014}, a learning algorithm based on contrastive learning was used to train a Reduced Boltzmann Machine implemented with spiking neurons. The authors propose a variant of STDP learning, with a free phase in which updates are performed according to the STDP rule, and a clamped phase in which the updates are anti-STDP. The authors show that the resulting algorithm approximates the traditional, non-spiking version of contrastive learning. Contrastive learning based on equilibrium propagation with spiking neurons was also studied on deeper models in \citep{oconnor2019}.

\subsection{Beyond Backprop Approximations} \label{sec:bp_alternatives:beyond}

While previous approaches leveraged bio-inspired mechanisms to construct backprop-approximating learning rules, other proposed methods have tackled the learning problem from different perspectives \citep{hinton2022, kohan2023, kulkarni2017, nokland2019, ma2020, pogodin2020, lukovsevivcius2009, lindsey2020, confavreux2020, miconi2018}.

Reservoir computing methods \citep{lukovsevivcius2009}, such as Liquid State Machines (LSM) \citep{maass2002} and Echo State Networks \citep{jaeger2001}, leverage a large \textit{reservoir} of recurrently connected (spiking) neurons, forming a highly nonlinear dynamical system. The reservoir is not trained; however, thanks to its size and dynamical behavior, it is able to nonlinearly map input stimuli to a high dimensional representation space, where linear separability is more easily achieved. A single readout layer is then trained to map the high dimensional representation to the desired output. A similar principle is adopted in Extreme Learning Machines (ELM) \citep{huang2006}, although this model uses feedforward connections only. Deep variants of reservoir architectures have also been proposed \citep{wang2016c, gallicchio2017}, and some works explored the possibility to tune the reservoir connectivity by STDP \citep{zhang2015a, wang2016c}. In \citep{disarli2020}, a gated variant of ESNs was proposed, to improve the model capacity of capturing long-term dependencies in the inputs, while \citep{cossu2021} showed promising results of reservoir computing models in continual learning scenarios \citep{hadsell2020}.

Layerwise training methods \citep{scardapane2020} provide a deep supervision signal on a layer-by-layer basis. A simple way to do so would be to attach a classifier on top of each hidden layer \citep{wang2015b, lee2015b, marquez2018}.  This layerwise training approach has proven effective in computer vision tasks \citep{belilovsky2019b}. This strategy does not remove the need for backpropagtion, but still reduces the backprop chains, alleviating gradient vanishing problems \citep{szegedy2015}. Other methods provide the supervision signal without a classifier on top of each layer, resorting instead to max-margin \citep{duan2021} or similarity matching \citep{kulkarni2017, nokland2019} criteria.

The Information Bottleneck (IB) principle \citep{tishby2000} has been proposed as an information theoretic approach to explain inference and generalization in learning systems. Let us consider a dependent variable $Y$ which has to be predicted from an independent variable $X$. The idea is that a deep representation $Z$ obtained by a learning agent, trying to discover the mapping between $X$ and $Y$, will maximize the information content the $Z$ carries about $Y$, while also compressing the irrelevant information from the input $X$ towards better generalization. The IB objective has the form:
\begin{equation}
    IB(Z) = I(Z, Y) - \beta I(Z, X)
\end{equation}
where $I(\cdot, \cdot)$ denotes the Mutual Information (MI) between two variables, and $\beta$ is a hyperparameter.
The IB approach has been applied for DNN training \citep{alemi2017}, using variational approximations for the MI, which would otherwise be difficult to estimate in practice. The approach leads to a generalized encoding-decoding architecture, where the encoder maps the input variable $X$ to a latent representation $Z$, and the decoder maps $Z$ to variable $Y$, in a possibly different target domain. Other approximations for the IB principle rely on simplified independence criteria, such as the Hilbert-Schmidts Information Criterion (HSIC)
\citep{chalk2016, duan2021}. These approaches leverage kernelized mappings to discover the (possibly nonlinear) dependence relationships between variables. Layerwise application of the HSCI IB principle for DNN training has been shown to be suitable for backprop-free training \citep{ma2020}, and to lead to three-factors Hebbian learning rules \citep{pogodin2020}.

Some recent works addressed the metalearning problem from a bio-inspired perspective. Before we have mentioned the idea of combining error-driven learning with Hebbian learning in order to combine task-specific and general knowledge. This idea was also shown to be effective in the context of transfer learning \citep{magotra2019, magotra2020} and metalearning \citep{miconi2018}, thanks to the transferability of unsupervised Hebbian features. In particular, in \citep{miconi2018} the \textit{differentiable plasticity} model was introduced, where a synapse is assumed to be composed of two parts: a backprop weight and a Hebbian weight. A weighting factor also determines the balance between the two contributions. The weight on a differentiable plasticity-based synapse can be written as:
\begin{equation}
    w = w_{BP} + \zeta \, w_{H}
\end{equation}
where $w_BP$ is the backprop part of the weight, $w_H$ is the Hebbian part, and $\zeta$ is the weighting factor.
The backprop part and the weighting constant are meta-learned, so that they learn to exploit the general information extracted from the Hebbian part for the specific task. The vanilla Hebbian plasticity model is used in this case, but still, the experimental analysis showed promising results on metalearning benchmarks. A related approach proposes Hebbian fast weights \citep{munkhdalai2018} in a melalearning context, i.e. Hebbian learning in the inner loop and backprop-based metalearning in the outer loop of the \textit{learning-to-learn} process. Other researchers proposed to leverage the learning-to-learn paradigm in order to automatically discover local synaptic plasticity rules with good performance on the given family of tasks \citep{bengio1992, metz2018, hsu2018, confavreux2020, lindsey2020, najarro2020}.

Recent approaches tackle the learning problem in a forward-only perspective, for example, the Signal Propagation algorithm \citep{kohan2023}, or the Forward-Forward approach \citep{hinton2022}. The idea is to that both the sample and the corresponding target can be fed together to the network, as two different parts of the same input. This idea is related to the multi-view learning approaches discussed in Section \ref{sec:plasticity}, where, in this case, one view is represented by the sample, and another view is represented by the target variable. Training can be performed by a contrastive procedure, by which true sample-target pairs (positives), and randomized sample-target pairings (negatives), should be mapped to different representations.
For example, in the recently proposed Forward-Forward (FF) algorithm \citep{hinton2022}, this is done by computing the sum of the squared outputs of a given layer. The result is then passed through a sigmoid nonlinearity, which associates a probability with the sample-target assignment. The layer is optimized to maximize this probability for positive sample-target pairs, and minimize it for negative pairs. It can be noticed that the proposed objective is related to the vanilla Hebbian plasticity rule (Eq. \ref{eq:hebbian_vanilla}), which indeed can be obtained as a gradient step on the sum of squared activations objective. A connection between Hebbian updates and corresponding objective functions is also highlighted in \citep{miconi2021}. Hence, this learning approach reduces to a Hebbian update in the positive phase, and an anti-Hebbian update in the negative phase, showing again how interesting connections may arise between various fields of exploration, under the lenses of biological inspiration.

\section{Concluding Remarks} \label{sec:concl}

In this survey, we discussed SNN models of neural computation, as biologically grounded alternatives to traditional DNNs. We highlighted the challenges related to SNN training, and then we moved to the discussion of novel solutions to address the learning problem, as alternatives to traditional backprop-based algorithms, both for spiking and for traditional networks.

A current limitation of SNNs is that their performance still lags behind that of traditional DNNs. In addition, SNN simulation is typically more complex and time-consuming on traditional hardware. 
However, SNNs are promising towards the realization of energy efficient DL, by means of biological \citep{ruaro2005, lagani2021a, kagan2022} or neuromorphic computing \citep{shrestha2022, huynh2022, schuman2022, zhu2020, roy2019b}.

Overall, biologically inspired computing models can offer many opportunities for exploration to researchers. In this perspective, BIDL represents an exciting and promising research area at the intersection between computer science and neuroscience. Given the breadth of topics embraced by this field, future challenges will foster the collaboration among researchers with the most variegated background, towards the goal developing both novel technologies and a deeper understanding of the mechanisms behind intelligence \citep{hbp}.

\begin{acks}
\noindent This work was partially supported by: \\
- Tuscany Health Ecosystem (THE) Project (CUP I53C22000780001), funded by the National Recovery and Resilience Plan (NRRP), within the NextGeneration Europe (NGEU) Program; \\
- Horizon Europe Research \& Innovation Programme under Grant agreement N. 101092612 (Social and hUman ceNtered XR - SUN project); \\
- AI4Media project, funded by the European Commission (H2020 - Contract n. 951911); \\
- INAROS (INtelligenza ARtificiale per il mOnitoraggio e Supporto agli anziani) project co-funded by Tuscany Region POR FSE CUP B53D21008060008.
-
\end{acks}

\small
\bibliographystyle{ACM-Reference-Format}
\bibliography{references}

\appendix

\section*{Appendix}

\section{Useful Tools}

In this Appendix, we briefly mention useful tools for experimentation with bio-inspired learning techniques.

Neurolab \citep{neurolab} is a Pytorch-based framework for deep learning experiments designed to provide easy management of experimental configurations and results, and integrating features that are commonly necessary, such as checkpointing and resuming of experiments, hyperparameter tuning, or handling of random number generators for reproducibility. GPU acceleration is also guaranteed thanks to the underlying primitives offered by Pytorch. The publicly available repository offers implementations of bio-inspired synaptic plasticity models in a convenient Hebbian convolutional module.

Simulators like Brian \citep{brian}, Nest \citep{nest}, or Pynn \citep{pynn}, offer environments for the simulation of spiking neuron models, based on the Python language. These packages are well suited for biologically accurate simulation of spiking neurons. On the other hand, Bindsnet \citep{bindsnet} and snnTorch \citep{snntorch} are other Pytorch-based packages for SNN simulation, which enable efficient simulation of large-scale architectures thanks to the GPU acceleration provided by Pytorch. This makes the latter packages better suited for the simulation of SNNs for DL contexts.

Finally, we mention the Brain-Score benchmark \citep{brainscore}, which provides measures to evaluate the degree of similarity of DNN architectures compared to biological primate brains, in terms of neural activity. The benchmark uses data about neural activity recording during the presentation of visual stimuli to monkeys, and evaluates the correlation between the activity of the artificial network and the biological data, in correspondence of the same stimuli. The work \citep{brainscore} also shows that task performance (in this case on ImageNet \citep{imagenet}) of various architectures, and the Brain-Score similarity with biological data are highly correlated.

\end{document}